\pdfoutput=1
\documentclass[11pt]{article}

\usepackage[]{acl}

\usepackage{times}
\usepackage{latexsym}

\usepackage[T1]{fontenc}

\usepackage[utf8]{inputenc}
\usepackage{graphics}

\usepackage{microtype}

\usepackage{microtype}
\usepackage{graphicx}
\usepackage{subfigure}
\usepackage{booktabs} % for professional tables

\usepackage{makecell}

\usepackage{CJKutf8}

\usepackage{hyperref}
\usepackage{url}
\usepackage{graphicx}
\usepackage{booktabs}
\usepackage{multirow, multicol}
\usepackage{diagbox}
\usepackage{listings}

\usepackage{tabularx}
\usepackage{xspace}
\usepackage{amsmath}
\usepackage{float}
\usepackage{xcolor}
\usepackage{tablefootnote}

\usepackage{pifont}
\usepackage{amssymb}

\usepackage{enumitem}

\newcommand{\chin}[1]{\begin{CJK*}{UTF8}{gbsn}#1\end{CJK*}}

\newcommand{\TODO}[1]{{\color{blue} TODO: #1}}

\newcommand{\multiwoz}{MultiWOZ\xspace}
\newcommand{\risawoz}{RiSAWOZ\xspace}
\newcommand{\xrisawoz}{X-RiSAWOZ\xspace}

\DeclareSymbolFont{extraup}{U}{zavm}{m}{n}
\DeclareMathSymbol{\varheart}{\mathalpha}{extraup}{86}
\DeclareMathSymbol{\vardiamond}{\mathalpha}{extraup}{87}

\title{Benchmarks Underestimate the Readiness of Multi-lingual Dialogue Agents}

\author{
\fontsize{11pt}{11pt}\selectfont
 \makecell{
 Andrew H. Lee$^1$ \; Sina J. Semnani$^1$ \; Galo Castillo-López$^2$ \; G\"{a}el de Chalendar$^2$ \\ Monojit Choudhury$^3$ \; Ashna Dua$^4$ \; Kapil Rajesh Kavitha$^4$ \; Sungkyun Kim$^5$ \\ Prashant Kodali$^4$ \; Ponnurangam Kumaraguru$^4$  \; Alexis Lombard$^2$ \; Mehrad Moradshahi$^1$ \\ Gihyun Park$^5$ \; Nasredine Semmar$^2$ \; Jiwon Seo$^5$ \;  Tianhao Shen$^6$ \\ Manish Shrivastava$^4$ \; Deyi Xiong$^6$ \; Monica S. Lam$^1$  
 }
 \vspace{2pt}
\\
\fontsize{7.3pt}{7.3pt}\selectfont
\makecell{
$^1$ Computer Science Department, Stanford University, Stanford, USA
$^2$ Universit\'{e} Paris-Saclay, CEA, List, Palaiseau, France \\
$^3$ Mohamed bin Zayed University of Artificial Intelligence
$^4$ International Institute of Information Technology, Hyderabad, India \\
$^5$ Computer Science Department, Hanyang University, Seoul, South Korea 
$^6$ College of Intelligence and Computing, Tianjin University, Tianjin, China 
}
}

\begin{document}
\maketitle

\begin{abstract}
Creating multilingual task-oriented dialogue (TOD) agents is challenging due to the high cost of training data acquisition. Following the research trend of improving training data efficiency, we show for the first time, that in-context learning is sufficient to tackle multilingual TOD.

To handle the challenging dialogue state tracking (DST) subtask, we break it down to simpler steps that are more compatible with in-context learning where only a handful of few-shot examples are used.
We test our approach on the multilingual TOD dataset \xrisawoz, which has 12 domains in Chinese, English, French, Korean, Hindi, and code-mixed Hindi-English. Our turn-by-turn DST accuracy on the 6 languages range from 55.6\% to 80.3\%, seemingly worse than the SOTA results from fine-tuned models that achieve from 60.7\% to 82.8\%; our BLEU scores in the response generation (RG) subtask are also significantly lower than SOTA.

However, after manual evaluation of the validation set, we find that by correcting gold label errors and improving dataset annotation schema, GPT-4 with our prompts can achieve (1) 89.6\%-96.8\% accuracy in DST, and (2) more than 99\% correct response generation across different languages. This leads us to conclude that current automatic metrics heavily underestimate the effectiveness of in-context learning.~\footnote{We will release our code upon publication.}
\end{abstract}
\section{Introduction}
Conversational agents for multi-domain task-oriented dialogue (TOD) require large datasets to train, and much effort has been put into creating and annotating them, each one having tens of thousands of hand-annotated dialogue turns~\cite{budzianowski-etal-2018-multiwoz, byrne-etal-2019-taskmaster, quan-etal-2020-risawoz, zhu-etal-2020-crosswoz}.
  
Extending TOD agents to more than one language, especially low-resource languages introduces additional cost and challenge in finding native annotators for the target language. Most approaches involve translating data from one language, which is still labor intensive~\cite{ding-etal-2022-globalwoz, li2021multi-domain}.
Zero and few-shot approaches that have been proposed often have a large performance gap with more data-intensive methods, and are either only tested in the \emph{monolingual} setting~\cite{zhao2022descriptiondriven, li-etal-2021-zero, campagna-etal-2020-zero, campagna-etal-2022-shot} or rely on machine translation from a large preexisting dataset~\cite{moradshahi-etal-2023-zero}.

Recent advances in LLMs have made possible an extremely data-efficient approach: in-context learning~\cite{brown-etal-2020-icl}.
To the best of our knowledge, \citet{ahuja-etal-2023-mega} is the only work that experiments with in-context learning for \emph{multilingual} TOD.
For {\em monolingual} multi-domain TOD, \citet{hudecek-dusek-2023-large} experiment on the English datasets MultiWOZ~\cite{ye-etal-2022-multiwoz} and SGD~\cite{sgd}, and \citet{chung-etal-2023-instructtods} only on MultiWOZ.
Other research on MultiWOZ only focuses on DST, just one of the four subtasks in TOD~\cite{pan2023preliminary, heck2023chatgpt, hu-etal-2022-context}.
As we show in this paper, evaluating in-context learning models with reference-based automatic metrics, as is common practice in the aforementioned papers, leads to a significant underestimation of their performance. This is due to 1) annotation issues like wrong gold labels and poor annotation schema, and 2) metric issues like BLEU score~\cite{papineni-etal-2002-bleu} penalizing stylistic differences in responses that are otherwise perfect.

Multi-domain DST is challenging for large language models (LLMs). First, the size of the schema grows with the number of domains, so without sufficient examples in context, LLMs do not learn the proper use of each slot. Similarly, some slots are enumerated type and can only take values from a finite set, which LLMs overlook if not instructed properly. Lastly, each dataset for TOD has a predefined annotation schema, e.g. when to include or exclude slots from previous turns. All these are challenging to learn by the very few examples that can be used in in-context learning.
To address these challenges in DST, we devise a multi-stage pipeline for in-context learning using hierarchical prompting~\cite{lo-etal-2023-hierarchical}.

% \iffalse
% \TODO{move elsewhere Although using automatic translation reduced human annotation efforts needed, the authors acknowledge that the quality for low-resource languages may be much lower, and as we show in our analysis (Section~\ref{sec:human-evaluation}), annotation quality can drop significantly during this process.} \TCM{how is annotation quality worse? remember we're not re-annotating the dataset; only possible mis-annotation is mismatch between entities and for that we have normalization step.}
% \fi

This paper's main contribution is to show, for the first time, that in-context learning is effective for multi-lingual, multi-domain TOD, provided that the language of interest is one that LLMs are good at. 

We propose a multi-stage LLM-pipeline that uses hierarchical pipeline with an LLM-based entity normalization for the DST subtask, along with simple prompts for all other TOD subtasks. We test our technique on the largest multilingual multi-domain dataset \xrisawoz~\cite{moradshahi-etal-2023-x} in six languages: Chinese, English, French, Hindi, Korean, and code-mixed English-Hindi, and all subtasks of TOD.

Our manual analysis of all of validation set for DST and 1000 agent responses per language for RG shows that 
with our approach, GPT-4 achieves (1) 89.6\%-96.8\% accuracy in DST, assuming corrected gold labels and improved dataset annotation schema, and (2) more than 99\% correct generated responses across different languages. 
This suggests that in-context learning can be used to create high-quality multilingual multi-domain TOD agents.

\section{Related Work}
\subsection{Multilingual TOD Datasets}
Most multilingual TOD datasets are created by translating an existing dataset to new languages. For example, 
GlobalWoZ, AllWOZ~\cite{zuo2021allwoz} and Multi2WOZ~\cite{hung-etal-2022-multi2woz} are the result of translating all or a subset of various versions of \multiwoz~\cite{zang-etal-2020-multiwoz, eric-etal-2020-multiwoz} to 20, 8 and 4 languages respectively.
BiToD~\cite{lin2021bitod} follows the less popular approach of synthesizing dialogues and annotations in multiple languages and paraphrasing them by crowdworkers.

Another recent multilingual dataset \xrisawoz~\cite{moradshahi-etal-2023-x} was translated from \risawoz~\cite{quan-etal-2020-risawoz}, which contains human-written dialogues spanning 12 domains and has the lowest annotation error rate among popular TOD datasets~\cite{moradshahi-etal-2023-contextual}.
\citet{moradshahi-etal-2023-x} first manually translated the validation and test sets of \risawoz from Chinese to English. Then for other target languages, they translated from English, through a round of machine translation and human post-editing.
We opt to use this dataset for our evaluation because, contrary to previous datasets, it covers more domains, has a lower error rate, covers all TOD subtasks, and includes a mid-resource and a code-mixed language.
In addition, \xrisawoz is more suitable for evaluating LLM-based agents since it most likely has not leaked to LLM pre-training data yet, as older datasets like \multiwoz have~\cite{balloccu2024leak}.

\subsection{In-Context Learning for TOD}
In-context learning for TOD is a relatively new area of research, and virtually all existing work focuses on English, and mostly only the subtask of DST~\cite{pan2023preliminary, heck2023chatgpt, hu-etal-2022-context}.
The multilingual setting has been practically unexplored, with the exception of \citet{ahuja2023megaverse}. They use a straightforward prompt that uses one prompt per subtask, and provides a few input-output examples to the model. However, we posit that this simple approach does not handle the large schema in the multi-domain DST well, and suffers from the \emph{entity normalization} problem where some generated slot values might not exist in the ontology~\cite{ye-etal-2022-multiwoz}.
They report their results on \xrisawoz with different LLMs, and we compare with their strongest model based on GPT-4.

Another proposed improvement to this simple prompting approach for DST is to use a proxy belief state in natural language~\cite{chung-etal-2023-instructtods} or SQL~\citep{hu-etal-2022-context}, which are easier for LLMs to work with. However, none of these methods have been evaluated in the multilingual setting.
\section{Task-Oriented Dialogue Subtasks}
\begin{figure}
    \centering
    \includegraphics[width=1.0\linewidth]{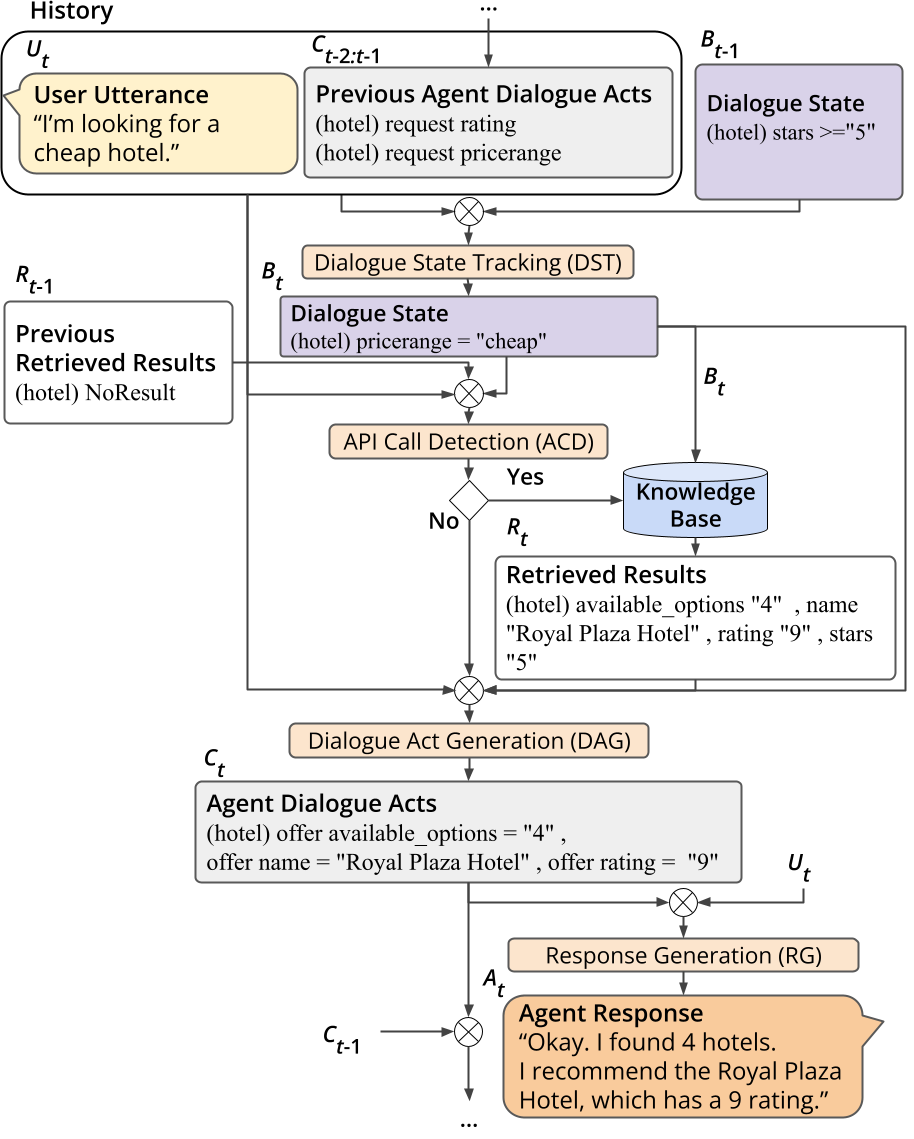}
    \caption{Subtasks of TOD and their connection with each other. Figure adapted from~\citet{moradshahi-etal-2023-zero}.}
    \label{fig:todsub}
\end{figure}

%\TODO{this section is too similar to \xrisawoz} \TCM{yeah both the text and figure needs adaptation. probably can chatgpt it :)}
In a task-oriented dialogue, a user interacts with an agent to achieve specific goals such as requesting recommendations or making bookings. While there are various designs on the subtasks that the agent has to perform in each turn, we follow the design of \citet{moradshahi-etal-2023-zero, moradshahi-etal-2023-x} to be consistent with X-RiSAWOZ. Figure~\ref{fig:todsub} illustrates the subtasks performed by the agent in each turn given the user utterance.

\begin{enumerate}
    \item \textit{Dialogue State Tracking (DST)}: Update the agent's dialogue state based on the user utterance at the current turn, considering the previous dialogue state and the last two dialogue acts.
    \item \textit{API Call Detection (ACD)}: Determine if querying the database via an API call is necessary based on the dialogue state, the last two dialogue acts and the current user utterance.
    \item \textit{Dialogue Act Generation (DAG)}: Generate the agent dialogue act based on the current user utterance, the dialogue state, the outcome of any API calls, and the last two dialogue acts generated.
    \item \textit{Response Generation (RG)}: Formulate the agent's dialogue act as a natural language response to user. 
\end{enumerate}
In our experiment, we evaluate LLMs on their performance on these four subtasks on \xrisawoz.

\section{In-Context Learning for Multidomain Multilingual TOD}

We prompt LLMs such that they can learn the TOD subtasks described in the previous section in the few-shot setting.

\subsection{In-Context Learning for DST}
DST is the most challenging among all the subtasks, as the dialogue state must use slots from the correct domain in the schema, and needs to match the slot values in the very large ontology. 
We use hierarchical prompting for the DST subtask to enforce schema and ontology. We address these challenges  with a pipeline of three stages, as summarized in Figure~\ref{fig:hierprompt}. The prompt for each module can be found in Appendix~\ref{section:prompts}.

% \iffalse
% For instance, in the following example, the correct label should not include the slot ``title'' because although the agent recommends this movie, it is neither a request the user has initiated or has confirmed.

% \begin{lstlisting}[basicstyle=\small]
%     DST: <state> ( tv ) production_country_or_area equal_to " Japanese TV show " , star equal_to " Jun Matsumoto " <endofstate> <history> AGENT_ACTS: ( tv ) recommend title equal_to " Lucky Seven " USER: Well, that seems like a good choice, but can you tell me its Douban rating? <endofhistory>
% \end{lstlisting}
% \fi

\begin{figure}
    \centering
    \includegraphics[width=1\linewidth, trim=20 210 30 70]{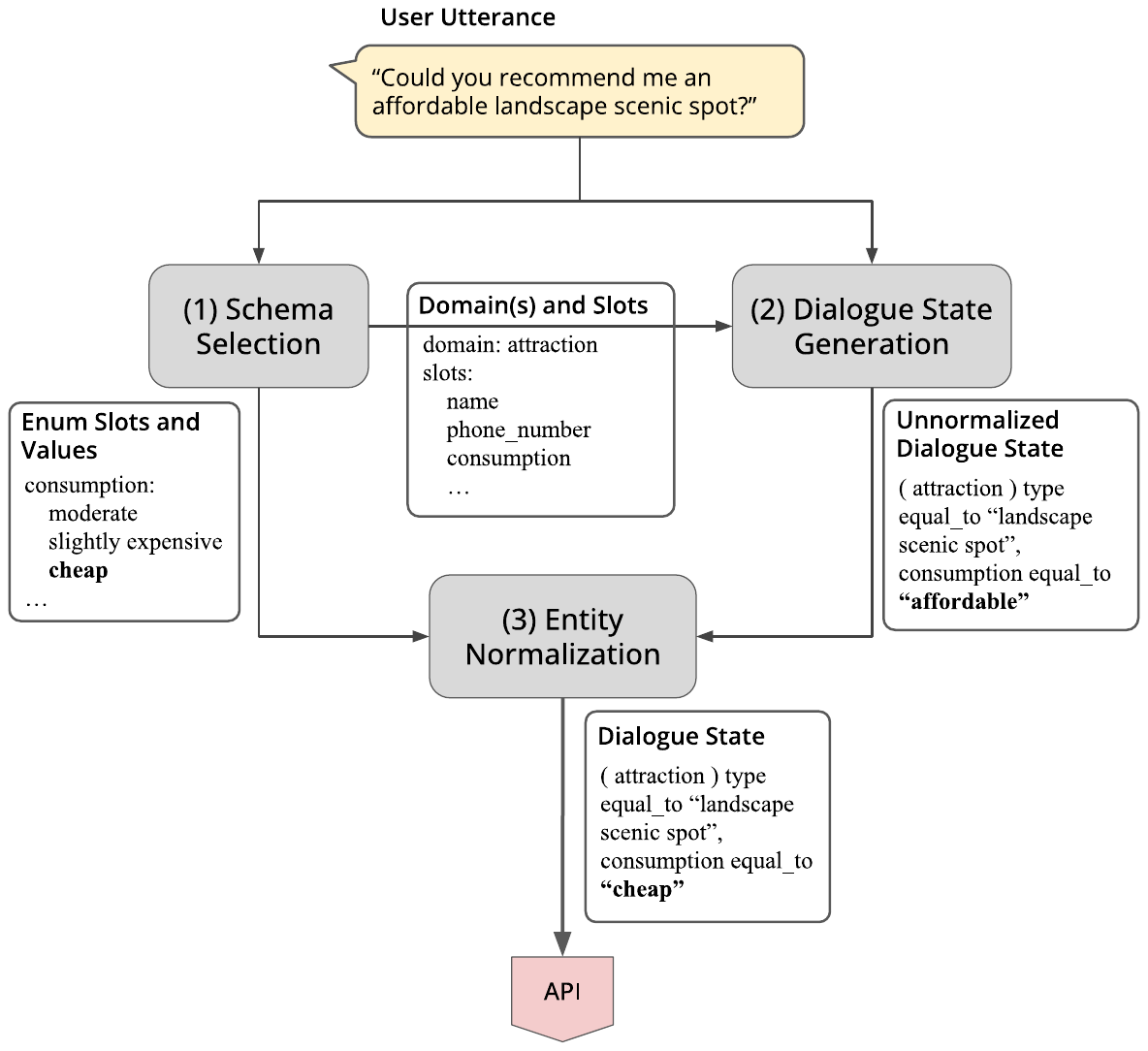}
    
    \caption{Hierarchical prompting for dialogue state tracking with LLMs.}
    \label{fig:hierprompt}
\end{figure}
%Previous works like Megaverse \tocite{Ahuja et al} and InstructTODS \tocite{Chung et al} prompt the DST subtask similarly to the other subtasks above. However, this leads to poor performance especially with the presence of multiple domains and \tocite{Chung et al} concludes ``LLMs do not solve multi-domain DST problems''. There are several reasons that make DST a relatively harder subtask. First, there is a fixed set of slots that are relevant to each domain; so without sufficient examples in context, LLMs will either leave out necessary slots or include irrelevant ones. Also, some slots have enumerated type for the value and LLMs often predict semantically similar values that do not adhere to this. Lastly, the DST subtask has non-trivial convention in what slots to include in the label. \TODO{Andrew: should I elaborate on what the convention is? Sina: After the moving this to intro, you can elaborate and include some examples here.}

\subsubsection{Schema Selection}
First, from the input, we use the LLM to determine one or more of the 12 domains that are mentioned so as to retrieve the relevant schemas.  The prompt includes i) instruction for the domain classification; ii) few-shot examples of DST data with relevant domains; and iii) the input example to extract domains from.

\subsubsection{Dialogue State Generation Using Schema}
Second, we instruct the LLM to predict the dialogue state using only the candidate slots in the given schemas. The challenge here is to ensure that the predicted dialogue state neither omits a necessary slot nor includes extraneous slots. \citet{hu-etal-2022-context} suggest a rather simple but effective solution of schema prompting in a similar situation - they concatenate SQL tables for all domains, each table containing a row of slot names followed by example values.
Applying their solution, we concatenate the schema for the domain as well as several few-shot examples of correct dialogue states using these slots. We further prepend these with a natural language instruction describing the subtask.

\subsubsection{Entity Normalization}
Third, we post-process slots of enumerated type in the database so that they match the values in the ontology. For instance, ``type'' in the TV domain can only take on values among ``comedy'', ``crime'', ``action'', etc. Originally, \citet{moradshahi-etal-2023-x} used deterministic dictionary-based entity mapping between each source entity and all possible translations in the target language to ensure the predictions match the ontology for their respective domains. Such a dictionary was built on-the-fly by translating different entities from sentences using nucleus sampling \cite{holtzman2020curious} with different temperature parameter. This approach still has limitations with respect to diversity of generated values as a dictionary made from sampling cannot cover indefinitely many possible mappings.

Instead, we instruct the LLM to normalize all slots of enumerated types by \emph{classifying} them into one of the allowed values. Similar to the other two steps, we include a natural language instruction for the prompt as well as few-shot examples. In Section~\ref{section:automatic-evaluation} we conduct an ablation study to show the effectiveness of our LLM-based entity normalization.

\subsection{LLM Prompting for ACD, DAG and RG}
For the ACD, DAG and RG subtasks, we use a direct few-shot prompting approach similar to \citet{ahuja2023megaverse}. The prompt for these three subtasks include a natural language instruction for the subtask, few-shot  examples of the subtask's inputs and outputs, and the test example drawn from the validation set of \xrisawoz, that LLM has to predict on. 
The few-shot examples are selected to be in the same language as the test example.

\subsection{Multilingual TOD}
Expanding any TOD dataset to new languages in accordance with a fine-tuning approach is very costly. We need bilingual speakers for source and target languages to either manually translate or edit the machine translations on tens of thousands of turns. We also need them to build the ontology and mappings for each language to ensure normalization.

With our setting, we are no longer relying on translations: all we need is a dozen of few-shot examples in the target language. We just need to make sure that these few-shot examples cover all domains to ensure generalization. One caveat is that the LLMs have to be proficient enough in the target language in order to achieve a high performance.
\section{Experiments}
\subsection{Dataset}
We test our approach on all six different languages that are covered by \xrisawoz, i.e. Chinese, English, French, Hindi, Korean, and English-Hindi mixed code. All few-shot examples used in our various prompts are obtained from the \textit{few-shot} split of \xrisawoz. The validation set contains 8,116 dialogue turns and 148 different slots.

\subsection{Models}
We experiment with two LLMs: GPT-4 (\texttt{gpt-4-1106}) and GPT-3.5 (\texttt{gpt-3.5-turbo-0301}) accessed via the OpenAI API. We use greedy decoding for all experiments, therefore, we report the results of a single run. For the baseline, we use the current state of the art (SOTA) on the dataset from \citet{moradshahi-etal-2023-x}, which fine-tunes models on ~134K automatically translated, and 1318 manually translated dialogue turns. mBART50~\cite{tang-etal-2021-multilingual} is used for all languages, except for Korean which uses M2M100~\cite{fan2020englishcentric}.

\subsection{Evaluation}
%\TODO{We only carry out manual analysis on a sample of the results from the GPT-4 model.}
We use the evaluation scripts from \citet{moradshahi-etal-2023-x} to perform 
turn-by-turn evaluation where we use the ``gold'' input of each turn, assuming the previous turns progressed perfectly.
We report the LLM performance on each subtask. 
For the DST, we use exact match (EM) accuracy.
For ACD, the prediction is a binary label and we use classification accuracy. The gold answer for DAG is one of a fixed number of possible acts: for instance, given a user request for restaurant recommendation in an area, the agent can either recommend one of the places from the query result or ask follow-up questions to narrow down. Therefore for DAG, while we report the EM for reference against fine-tuned model, we note that this figure does not fully capture the performance on the subtask. For example, \citet{moradshahi-etal-2023-x} report that 38\% of the EM ``errors'' result from multiple correct dialogue acts.
%\TODO{What are the possible acts, give some examples.}
The metrics for RG is BLEU score~\cite{papineni-etal-2002-bleu} which measures the overlap of the natural language response and the gold label. 

\begin{table*}[ht]
\centering
\scalebox{0.85}{
\begin{tabular}{l l c c c c c}
 \hline
& Model & DST Acc. & API Acc. & DA Acc. & RG BLEU & RG Avg. Length\\
 \hline
 
       & GPT-4 & 80.3 & 96.0 & 63.6 & 29.0 & 19.56 \\
Chinese   & GPT-3.5 & 58.4 & 87.6 & 59.2 & 23.1 & 21.38\\
       & Fine-tuned SOTA  & \textbf{82.8} & \textbf{98.6} & \textbf{77.3} & \textbf{38.9} & 11.29 \\
       
 \hline
 
       & GPT-4 & 74.9 & 96.0 & 61.2 & 30.6 & 13.39 \\
English   & GPT-3.5 & 55.6 & 88.4 & 42.4 & 25.5 & 15.31 \\
       & Fine-tuned SOTA & \textbf{84.6} & \textbf{98.7} & \textbf{69.4} & \textbf{46.4} & 9.04 \\
 \hline
 
       & GPT-4 & 66.5 & 95.2 & 45.2 & 34.5  & 14.67\\
French   & GPT-3.5 & 48.0 & 86.8 & 40.4 & 26.6 & 17.92\\
       & Fine-tuned SOTA & \textbf{73.1} & \textbf{99.1} & \textbf{61.1} & \textbf{42.2} & 10.87\\
 \hline
 
       & GPT-4 & 55.6 & 92.3 & 41.6 & 25.2 & 12.55\\
Hindi   & GPT-3.5 & 31.6 & 84.4 & 37.6 & 15.9 & 18.69\\
       & Fine-tuned SOTA & \textbf{75.2} & \textbf{99.1} & \textbf{59.0} & \textbf{38.4} & 10.03\\
\hline

       & GPT-4 & 66.2 & 96.4 & \textbf{58.8} & 23.6 & 7.98\\
Korean   & GPT-3.5 & 46.8 & 84.8 & 45.6 & 15.9 & 9.82\\
       & Fine-tuned SOTA & \textbf{71.2} & \textbf{98.6} & 53.5 & \textbf{34.9} & 5.84\\

\hline

       & GPT-4 & \textbf{62.2} & 96.0 & 33.6 & 22.6 & 12.32\\
English-Hindi   & GPT-3.5 & 44.0 & 84.8 & 24.0 & 13.7 & 17.67\\
       & Fine-tuned SOTA & 60.7 & \textbf{98.7} & \textbf{38.0} & \textbf{26.8} & 9.96\\

 \hline
\end{tabular}
}
\caption{Comparison of automated metrics with the state-of-the-art few-shot technique and our in-context learning on the validation set of \xrisawoz, obtained by feeding the gold input for each subtask in each turn. The best number for each language is in bold. Fine-tuned SOTA is the best few-shot model from \citet{moradshahi-etal-2023-x}.}
\label{table:tbtresults}
\end{table*}

%Table 2
\begin{table*}[ht]
\centering
\scalebox{0.85}{
\begin{tabular}{l c c c c c c}
 \hline
 & Chinese & English & French & Hindi & Korean & English-Hindi\\
 \hline
        Our (GPT-4) & \textbf{80.3} & \textbf{74.9} & \textbf{66.5} & \textbf{55.6} & \textbf{66.2} & \textbf{62.2}\\ 
        \quad w/o any normalization & 77.1 & 72.7 & 61.6 & 34.8 & 61.3 & 37.7\\
        \quad w/ dictionary-based normalization  & n/a & 74.6 & 66.4 & 46.0 & 63.4 & 40.4 \\
        Naive prompting & 60.8 & 62.5 & 52.9 & 30.0 & 44.4 & 33.2\\
 \hline
\end{tabular}
}
\caption{Ablation on different stages of our proposed prompting technique for DST. Numbers are exact match with the gold labels. Dictionary-based post-processing is from \citet{moradshahi-etal-2023-x}. Naive prompting is our implementation of~\citet{ahuja-etal-2023-mega}.}
\label{table:ablations}
\end{table*}

\section{Experimental Results}

Our evaluation of LLMs consists of two parts: (1) automated evaluation using the metrics defined in the original X-RiSAWOZ dataset, and (2) human evaluation, which we argue is critical for the proper assessment of LLM-based systems. 

%We don't want to highlight the 10% only testing and talk about mismatches here.  We evaluate LLMs on the validation set of X-RiSAWOZ with turn-by-turn metrics. We also provide the results on approximately 10\% of the test set with end-to-end metrics used for evaluation in \cite{moradshahi-etal-2023-x}. Along with these automated metrics, we conduct human evaluation on all mismatches for DST and RG to compare the quality of GPT-4 predictions against the labels.

\subsection{Evaluation with Automated Metrics}
\label{section:automatic-evaluation}
We report the turn-by-turn metrics for LLM-based models and fine-tuned baselines in Table~\ref{table:tbtresults}. 

Fine-tuned baselines outperform LLM prompting across all TOD subtasks according to the automated metrics. 
Our turn-by-turn DST accuracy on the 6 languages range from 55.6\% to 80.3\%, slightly lower than the SOTA results from fine-tuned models that achieve from 60.7\% to 82.8\%; our BLEU scores in response generation (RG) range from 22.6 to 34.5, much lower than the SOTA 26.8 to 46.4. 
The fine-tuned models are better at learning all the different features (e.g. word choice, sentence length, syntax etc.) of the label. For DST, they include dialogue state annotation schema and values for enumerated types. For RG, the fine-tuned models tend to generate responses of similar style and length with the labels, leading to a better BLEU score. LLMs are penalized for being 24\% to 73\% more verbose.

% Table 4
\begin{table*}[ht]
\centering
\scalebox{0.85}{
\begin{tabular}{l | rr | rr | rr | rr | rr | rr}
 \hline
  & \multicolumn{2}{c}{Chinese} & \multicolumn{2}{c}{English} & \multicolumn{2}{c}{French} & \multicolumn{2}{c}{Hindi} & \multicolumn{2}{c}{Korean}& \multicolumn{2}{c}{English-Hindi}\\
 \hline
 Cause of mismatch & \% & Acc. & \% & Acc. & \% & Acc. & \% & Acc. & \% & Acc. & \% & Acc.\\
 \hline
         & & 80.3 & & 74.9 & & 66.5 & & 55.6 & & 66.2 & & 62.2\\
        Multiple Correct Answers& 0.0 & 80.3 & 4.4 & 79.3 & 4.5 & 71.0 & 2.5 & 58.1 & 2.4 & 68.6 & 5.2 & 67.4\\
       Poor Gold Label & 4.0 & 84.3 & 9.3 & 88.6 & 12.8 & 83.8 & 20.1 & 78.2 & 15.2 & 83.8 & 16.5 & 83.9\\
       Poor Annotation Schema & 11.9 & \textbf{96.2} & 8.3 & \textbf{96.8} & 6.5 & \textbf{90.3} & 11.4 & \textbf{89.6} & 9.8 & \textbf{93.6} & 8.5 & \textbf{92.4}\\
   Error & 3.8 & & 8.3 & & 9.7 & & 10.4 & & 6.4 & & 7.6 &\\
       \quad Domain & 0.0 & & 0.3 & & 0.1 && 1.6 & & 0.3 & & 0.5\\
       \quad Slot & 3.1 & & 1.7 & & 3.3 && 2.2 & & 2.2 & & 3.3 & \\
       \quad Slot Value & 0.3 & & 0.4 & & 1.0 && 4.8 & & 2.0 & & 1.3 &\\
       \quad\quad Post-processing & 0.3 & & 0.8 & & 1.7 && 1.6 & & 1.8 & & 2.3 &\\
 \hline
\end{tabular}
}
\caption{Turn-by-turn DST error Analysis for our GPT-4-based hierarchical prompting on the \xrisawoz entire validation set. Acc. is the cumulative exact match accuracy if the dataset issue in that row is fixed. Numbers in bold are the obtainable accuracy after all issues are fixed.}
\label{table:dstana}
\end{table*}

%Table 5
\begin{table*}[ht]
\centering
\scalebox{0.85}{
\begin{tabular}{l r r r r r r}
 \hline
 & Chinese & English & French & Hindi & Korean & English-Hindi\\
 \hline
        Poor Gold DA Label & 7 & 9 & 13 & 2 & 6&13\\ 
        Poor Gold RG Label & 0 & 6 & 8 & 7 & 3&8\\
        Error & 5 & 4 & 8 & 5 & 5&5\\
        \quad Wrong Language & 0 & 0 & 0 & 0 & 3&0\\
        \quad Wrong Act & 0 & 1 & 0 & 1 & 0 &0\\
        \quad Wrong Entity & 2 & 3 & 7 & 3 & 0 &4\\
        \quad Wrong Semantics & 3 & 0 & 1 & 1 & 2&1\\
        \hline
        {\bf Total} & 12 & 19 & 29 & 14 & 14&26\\
 \hline
\end{tabular}
}
\caption{Turn-by-turn RG error analysis for our GPT-4-based hierarchical prompting on the \xrisawoz validation set. 1000 samples are analyzed per language.}
\label{table:rg_analysis}
\end{table*}
Table~\ref{table:ablations} shows the effectiveness of our strategy compared to naive prompting of \citet{ahuja-etal-2023-mega}; the difference ranges from 12.4 to 29.0\%. Our LLM-based normalization improves the accuracy by 20.8\% and 24.5\% for Hindi and English-Hindi, and between 2.2\% and 4.9\% for all other languages. We also compare our entity normalization with the dictionary-based approach by \citet{moradshahi-etal-2023-x}. A dictionary mapping common words to the normalized enumerated values were constructed during the translation process from the original Chinese dataset to other languages, hence there is no dictionary for Chinese. The LLM-approach requires no manual effort, and is significantly better for Hindi and English-Hindi by 9.6\% and 22.2\%, respectively. 
In summary, even though our technique improves over naive prompting, it still trails behind the SOTA according to the automatic comparison with gold labels.

\subsection{Human Evaluation}
\label{sec:human-evaluation}
Our analysis on the mismatches between LLM outputs and gold labels reveals that our method performs much better than indicated by the measured result. The manual analysis for DST was performed on the entire validation data set. 
Our analysis identifies three different categories of mismatches, as shown in Table~\ref{table:dstana}:

{\bf Annotation problem 1: multiple correct answers}. For example, \textit{``pas trop cher''} (not too expensive) and \textit{``bon marché''} (cheap) are two canonical values for an enumerated type, but they have similar meanings. So we need to accept both values as acceptable answers. This accounts for up to 5.2\% of the data set. 

{\bf Annotation problem 2: poor gold labels}. Errors include gold label having an incorrect dialogue state, incorrect canonical slot values, or not following the annotation schema. The gold label error rate is just 4.0\% for Chinese, the language of the original RiSAWOZ dataset, 8.3\% for English, which was hand-translated, and 12.8--19.7\% for the rest, which were automatically translated and hand-corrected. In other words, each translation process adds to the inaccuracy of the annotations. With in-context learning, LLMs are not exposed to such errors, and can generate correct answers from 77.8\% to 88.6\% across languages. 

{\bf Poor annotation schema}. We attribute a significant class of mismatches to the schema design of the ``recommend'' and ``inform'' acts in RiSAWOZ. Only slots that have been agreed upon or confirmed by the user shall be updated in the dialogue state. This design is problematic because the user may refer to the recommended value without confirming it.

Consider the example with the Agent Act being `(TV) recommend title equal\_to "Lucky Seven" ' and the user response being `Well, that seems like a good choice, but can you tell me its Douban rating?' According to this schema design, the agent's suggested ``Lucky Seven'' title should not be included in the ``title'' slot because it is neither a request the user has initiated or has confirmed. Note that not capturing this information in the formal dialogue state would make it {\em impossible} for the agent to actually answer the user query correctly.

Even though we tried to use prompt instructions and few-shot examples to teach this unusual annotation schema in our prompt, the LLM is not conducive to adhering to such a design and would  include the {\em necessary} slot value.
A better design would be to allow the indication of proposed slot values, then LLMs would likely learn the schema well. We estimate the accuracy in that case to be between 89.6\% and 96.8\% across languages.

{\bf LLM errors}. The true error rates range only from 3.2-11.0\% across different languages. They can be caused by wrong domain classification (often in the case of multi-domain inputs), wrong update of a new slot or the value of an existing slot, or failing to properly normalize entities for enumerated slots. 

%\TCM{besides the "Poor gold label" I'm confused how other analysis is supporting this conclusion that LLMs do a better job at annotating a dataset than humans.}

%Without this analysis, it may appear that more research is necessary to improve upon multilingual dialogue state tracking before such agents are useful in practice.  We show, in fact, that we have reached the limit of training models with manually annotated data and LLMs are all you need for multilingual slot-based DST. Furthermore, we need to examine our conventions to ensure that they are reasonable and easy for LLMs to work with. 

{\bf Response Generation}. In Table~\ref{table:rg_analysis}, we perform a similar breakdown for RG on 20\% of the validation set (1000 turns per language). Despite the low BLEU scores, manual analysis indicates that there are only 12-29 incorrect responses generated in 1000 turns. Poor gold labels in dialogue acts (DA) which are the input to the RG subtask, and poor gold RG labels account for the majority issues; only 4-8 of cases are true errors.

\subsection{Language-Specific Observations}
In the following, we highlight the unique observations made for each of the languages. These discussions highlight the shortcomings of hand-annotated data and suggest ways to improve LLM-based TOD systems.

\subsubsection{Chinese}
% \TODO{I've listed my observations during annotation here. Please take a look and make revision as you see fit. (Tianhao)}
% My main observations (Tianhao)
% DST:
% 1. Most slot value errors (over 90%) are because the model output values that are not in canonicalized form in ontology, while other 10% are semantically wrong slot values.
% 2. Most slot errors (over 80%) are because of convention mismatch.
% 3. Most erroneous ground truth annotations are because they inherit the (domain, slot, value) triplets from the "recommend" action.
% 4. GPT-4 tends to extract wrong dialogue states from a question. For example, if a user asks: "Is there any subway directly to ABC restaurant?", GPT-4 tends to add "metro_station": "yes" to dialog states.
% RG: 
% 1. 90% just paraphrase, even no notable difference in user-friendliness
% 2. ~10 of GPT-4 responses are incorrect.
% 3. GPT-4 tends to repeat what have been told by user in last turn so it is wordy and unnatural sometimes.
For DST, the Chinese dataset has the least amount of annotation error, but still suffers a significant failure rate due to the poor schema design for the ``recommend'' and ``inform'' dialogue acts. Of the true errors, the inaccuracies stem from normalization and annotation schema mismatches. 

For the RG subtask, we observe that GPT-4 can be unnecessarily verbose, without offering a noticeable enhancement in user-friendliness. For example, the labeled response to ``\chin{坐地铁能直达吗}'' (``Can I get there directly by subway?'') is ``\chin{可以的呢}'' (``Yes, you can.''), whereas GPT-4's answer is ``\chin{是的，你可以乘坐地铁直达这个景点}'' (``Yes, you can directly reach the destination by subway.'').

\subsubsection{English}
English has the lowest true error rate, 3.2\% and 0.12\% for DST and RG respectively. Most DST errors are due to incorrect updates when new slots are introduced in the user utterance. For RG, the three counts of wrong entity were all incorrect entity detection when the user utterance and previous dialogue acts did not make the subject clear.

\subsubsection{French}
Our analysis on the DST task suggests that many errors stem from a poorly designed ontology. For instance, the expression \textit{``pas trop cher (not so expensive)''} is interpreted as \textit{``moderate''} or \textit{``low''} prices by GPT-4 and mBART, respectively. As a result, mismatches occur when the former infers \textit{``moderate''} and the latter infers \textit{``low''}. 

Another major source of errors is multi-domain settings where the user asks for a recommendation within an area (e.g. hotel) by using the expression \textit{``a proximité (nearby)''}, when the area was mentioned as a part of another domain (e.g. attraction in Sichuan). GPT-4 occasionally does not infer the area for the new domain correctly in these cases.

For the RG subtask, we generally find that GPT-4 tends to produce more user-friendly responses, while mBART and hence the label tends to be more concise. We note that sometimes GPT-4's verbosity may not be desirable. 
%Nevertheless, such an analysis requires additional investigation.

\subsubsection{Korean}
We find that high proportion of errors with the label comes from labels themselves failing to follow the DST annotation schema. On the other hand, post-processing failure is the result of the language trait specific to Korean: \textit{jo-sa}, which is used to delimit phrases in Korean, interferes with the post-processing step. This is an instance also observed with dictionary-based normalization in \citet{moradshahi-etal-2023-x}.

For the RG subtask, we found that the responses from GPT-4 tend to be less ambiguous and more user-friendly. Our analysis further reveals that some of asymmetry in generated responses can be attributed to the subtle nuances of Korean language: for instance, when inquiring about restaurant ratings, Koreans are more inclined to ask ``Is the rating high?'' rather than ``What is the rating?'' While such user utterances were translated accordingly in the creation of X-RiSAWOZ, the label would simply provide the rating as they were translated separately from the user utterance. GPT-4 generates better responses in these situations.

\subsubsection{Hindi}
%most of the errors are caused by the incorrect labels, followed by poor convention. 

For Hindi, a major reason EM metric can be problematic is that there are multiple ways of spelling the same Hindi word. With this, neither a dictionary-based nor, to a lesser extent, an LLM-based normalization can cover all the bases, resulting in the highest true error rate among all languages. Also, both the label and LLM predictions fail to adhere to annotation schema in many cases.
%I need an example. 
%We found the common error lies in the model being overzealous in copying the slot/slot-values in the input. It also often includes erroneously information present in 
%contains information ``AGENT\_ACTS'', or ``Prev\_ACTS''.
%In case of incorrect slot, slot-values, in most cases, model output contained additional slots/values copied from the input, indicating the model was either over-zealous in copying the slot/slot-values present in the model input. Further, model prediction often contained information present in ``AGENT\_ACTS'', or ``Prev\_ACTS'', as slot or slot values. 

% \iffalse
% LET's RESOLVE THIS AFTER SUBMISSION. I don't think this is the case - the metrics I used ensures that ordering itself does not cause errors. Those examples should have other errors within them.
% \TODO{This is surprising that we didn't normalize the slot order-Monica} 
% \TCM{slot orders were normalized in \xrisawoz. are we not doing that in this paper?} Additionally, we also noticed that the order in which slots were mentioned would be different across model prediction and gold label, while they were technically correct, Exact Match criteria would mark such cases as error. 
% \fi 

For RG, we observe that often, GPT-4's verbosity helps with clarity, fluency and user-friendliness. But sometimes, it gives an expanded version of the label response without improving clarity.

\subsubsection{English-Hindi Mixed Code}
While the code-mixed English-Hindi~(en-hi) inherits a lot of similar errors from Hindi, its accuracy is better than Hindi while worse compared to English. The most common error is slot error - either including extraneous, incorrect slots or omitting a necessary slot in the dialogue state. For RG, we find that in a lot of the cases, GPT-4 generates unnecessarily verbose responses which affect clarity. 

%For en-hi DST, the distribution of errors is very similar en-hi, with improper gold label, incorrect convention, and multiple correct DSTs accounting for most of the errors. 

\subsection{Discussion}

We observe that once we account for the annotation issues of multiple correct answers and poor gold labels, the accuracy for all the different languages is much more consistent, ranging from 77.8\% to 88.6\%. There is only an 11\% difference rather than 24.7\% observed using automated evaluation. The variance is caused by the varying error rates in annotation across the languages. We note that annotation errors in manually annotated TOD datasets have been well documented~\cite{eric-etal-2020-multiwoz, zang-etal-2020-multiwoz, han2021multiwoz, ye-etal-2022-multiwoz, campagna-etal-2022-shot}.
% Note that RiSAWOZ is the most accurate and largest multi-domain TOD, and \xrisawoz is a 6-language dataset based on RiSAWOZ, we cannot find other comparable datasets to experiment with.

In-context learning only needs a few correctly labeled examples, and thus is not subjected to errors of manual annotation.  
Our results show that, assuming a reasonable annotation schema, there is little room for improvement in DST or RG on multilingual and multi-domain datasets like \xrisawoz. 
In view of the huge discrepancy observed between automated metrics and manual evaluation, further evaluation such as measuring our technique on the test set, without manual analysis, is not meaningful.
\section*{Limitations}
Given that our results rely on large language models, we expect the quality of our dialogue agent to be limited by the quality of the underlying LLM. Specifically, it is likely that there will be a significant drop in quality when this work is evaluated on truly low-resource languages, but unfortunately, we are not aware of any such TOD datasets.

Another limitation is that we experiment with \xrisawoz which uses slot-values, as opposed to more complex representations like SQL, or less common representations like TreeDST~\cite{cheng-etal-2020-conversational}, dataflow~\cite{dataflow} or ThingTalk~\cite{lam2022thingtalk} which LLMs might not be familiar with.
In addition, \xrisawoz has relatively limited inter-domain dependencies (e.g. complex ``parameter passing'' between APIs in different domains), which may make the task simpler than some real-world applications require.
Designing representation for TOD subtasks that are both more expressive and suitable for use with LLMs is a promising direction for future work.

Lastly, we acknowledge that our design and experiments were specific to X-RiSAWOZ to make generalization to multi-lingual TOD. While the hierarchical prompting design is specific to the TOD subtasks design behind X-RiSAWOZ, a similar method can be applied to test other designs too which we leave as future work. From our analysis, we observe that LLM predictions often surpass the labels for data sets relying on machine-translation and some fixed annotation schema due to translation errors and suboptimal cross-language alignments. Such shortcomings in the setup can be generalized to most of the existing multi-lingual ToD datasets, including X-SGD~\cite{sgd-x}.

% \newpage

\section*{Ethical Considerations}

No crowdsourcing or data collection is done in the process of writing this paper, and manual analysis is done by the authors.
We hope that this research can enable further task-oriented dialogue research in non-English languages and especially lower resource languages.

We do not anticipate harm resulting from the approaches we propose in this paper.

The dataset used in this paper was released by \citet{moradshahi-etal-2023-x} under a permissible license (BSD 3-Clause License), and does not include any personally identifiable information.
Upon publication, we will release our code to reproduce our results under Apache-2.0 license.

We did not conduct any experiments using local models, and only used LLMs via the OpenAI API. Thus, it is not feasible to estimate the total computation used. Nevertheless, we cached the results of the API calls wherever possible to reduce the number of API calls, and report that the total cost of our experiments was about \$1500.

%\iffalse
% no acks in submission
% \section*{Acknowledgements}
% We would like to thank Ruchi Jain for helping us validate the automatically translated Hindi dialogues.
% This work is supported in part by the National Science Foundation
% under Grant No.~1900638, the Alfred P. Sloan Foundation under Grant No.~G-2020-13938, the Verdant Foundation, Microsoft, KDDI, JPMorgan Chase, and the Stanford Human-Centered Artificial Intelligence (HAI) Institute.
% This work is also co-funded by the Natural Science Foundation of Xinjiang Uygur Autonomous Region (No. 2022D01D43), Zhejiang Lab (No. 2022KH0AB01). This project has also received funding from the European Union’s Horizon 2020 Research and Innovation Programme under Grant Agreement No. 101021797 (Starlight), and the European Union’s Horizon Europe research and innovation programme under grant agreement N° 101070192 (CORTEX²). 
% This work is also supported in part by Institute of Information \& communications Technology Planning \& Evaluation (IITP) grant funded by the Korea government(MSIT) (No.2020-0-01373 and IITP-2022-2021-0-01817).

%\fi

\bibliography{anthology,custom}

% move appendix to a new page so we can cut it out
% need to submit as a separate file
% \newpage
\appendix
\section{Appendix}
\subsection{LLM Prompts for DST}
\label{section:prompts}

We include all the prompts used for the in-context learning of the DST subtask.
The first step is extracting the relevant domains from the input. (Table~\ref{prompt:domain-selection})

We then list all the possible slots for the extracted domain(s) as well as few-shot examples from those domain(s) in the prompt so that LLM can predict dialogue state using the correct schema (Table~\ref{prompt:dst})

Lastly, we normalize slots that are enumerated type. We explicitly list all the enum fields and possible values for each domain predicted in the first step, then instruct the LLM to normalize the values by providing few-shot examples of normalization.
The expected output is the dialogue state with all slot values normalized (Table~\ref{prompt:normalization}).

\begin{table*}
\begin{lstlisting}[basicstyle=\ttfamily\small]
{# Instruction #}
Similar to the examples below, retrieve all the relevant domains from the following choices: 
movie, tv, attraction, retaurant, car, hotel, hospital, weather, flight, pc, train, class.

{# Few-shot example 1 (single domain) #}
<state>
    null
</state>
<history>
    User: Hi, can you recommend some fascinating Japanese TV shows?
</history>
Domain(s): tv

{# Few-shot example 2 (multi-domain) #}
<state>
    ( tv ) decade equal_to " 2010s " , production_country_or_area equal_to " Japanese TV show " , 
    type equal_to " suspenseful "
</state>
<history>
    Agent acts: ( tv ) inform Douban_score equal_to " 9.1 "
    Agent acts: ( tv ) inform director equal_to " Nobuhiro Doi "
    User: Thanks, could you recommend an Indian movie?
</history>
Domain(s): movie, tv

... {# More few-shot examples to cover at least one example per domain. #}

{# Input #}
<state>
    {{ B_t-1 }}             {# Dialogue state in the previous turn #}
</state>
<history>
    Agent acts: {{ C_t-2 }} {# Agent acts from two turns ago, if exists#}
    Agent acts: {{ C_t-1 }} {# Agent acts from the previous turn, if it exists #}
    User: {{ U_t }}         {# The last user utterance #}
</history>
Domain(s):
\end{lstlisting}
\caption{DST schema selection prompt. This prompt has 13 few-shot examples.}
\label{prompt:domain-selection}
\end{table*}

\begin{table*}
\begin{lstlisting}[basicstyle=\ttfamily\small]
{# Instruction #}
Similar to the examples below, generate a dialogue state using only the slots that the user 
has confirmed or agreed upon. A list of possible slots for each domain is included below.

{{ domain_name_1 }}: {{ list_of_domain_slots_1 }}

{# Few-shot example 1 of domain 1 #}
<state>
    null
</state>
<history>
    User: Hi, can you recommend some fascinating Japanese TV shows?
</history>
Output: ( tv ) production_country_or_area equal_to " Japanese TV show "

... {# More few-shot examples to cover appropriate usage of all domain slots. #}

... {# If more than one domain is chosen, we include examples for additional domains here. #}

{# Input #}
<state>
    {{ B_t-1 }}             {# Dialogue state in the previous turn #}
</state>
<history>
    Agent acts: {{ C_t-2 }} {# Agent acts from two turns ago, if exists#}
    Agent acts: {{ C_t-1 }} {# Agent acts from the previous turn, if it exists #}
    User: {{ U_t }}         {# The last user utterance #}
</history>
Output:
\end{lstlisting}
\caption{DST dialogue state generation prompt. This prompt has 7 few-shot examples per domain.}
\label{prompt:dst}
\end{table*}

\begin{table*}
\begin{lstlisting}[basicstyle=\ttfamily\small]
{# Instruction #}
Similar to the examples below, normalize any listed slots below that appear in the dialogue 
state to one of the possible values. Do not change the dialogue state if not necessary.

{{ domain_name_1 }}: {{ list_of_possible_slot_values_1 }} {# E.g. "tv" : 
"production_country_or_area" : ["Taiwan", China", "America", ...], 
"type" : ["romantic", "sci-fi", ...], ... #}

{# Few-shot example 1 #}
Input: ( tv ) production_country_or_area equal_to " United States " , 
type equal_to " science fiction TV show "
Normalized: ( tv ) production_country_or_area equal_to " America " , type equal_to " sci-fi "

... {# More few-shot examples. #}


{# Input #}
Input: {{ B_t }}  {# Dialogue state in the current turn, before entity normalization #}
Output:
\end{lstlisting}
\caption{DST entity normalization prompt. This prompt has 12 few-shot examples.}
\label{prompt:normalization}
\end{table*}

\end{document}